\pdfoutput=1

\documentclass[11pt]{article}

\usepackage[final]{coling}

\usepackage{times}
\usepackage{latexsym}

\usepackage{multirow,diagbox,makecell,amssymb,bm}
\usepackage{amsmath,amsthm}
\usepackage{makecell}
\usepackage{graphicx}
\usepackage{pifont}
\usepackage{algorithm}
\usepackage{algorithmic}
\usepackage{amsfonts}       
\usepackage{booktabs}       

\usepackage[T1]{fontenc}

\usepackage[utf8]{inputenc}

\usepackage{microtype}

\usepackage{inconsolata}

\usepackage{graphicx}

\title{Diversity-grounded Channel Prototypical Learning for Out-of-Distribution Intent Detection}

\author{
Bo Liu$^1$, Liming Zhan$^1$, Yujie Feng$^1$, Zexin Lu$^1$, Chengqiang Xie$^1$, \\ \bf Lei Xue$^2$, Albert Y.S. Lam$^3$, Xiao-Ming Wu$^1$\Thanks{~ Corresponding author.} \\ 
$^1$Department of Computing, The Hong Kong Polytechnic University, Hong Kong S.A.R. \\
$^2$School of Cyber Science and Technology, Sun Yat‐Sen University, Shenzhen, China \\
$^3$Fano Labs, Hong Kong S.A.R. \\
{bokelvin.liu@connect.polyu.edu.hk, csxmwu@comp.polyu.edu.hk}\\
}

\begin{document}
\maketitle
\begin{abstract}

In the realm of task-oriented dialogue systems, a robust intent detection mechanism must effectively handle malformed utterances encountered in real-world scenarios. 
This study presents a novel fine-tuning framework for large language models (LLMs) aimed at enhancing in-distribution (ID) intent classification and out-of-distribution (OOD) intent detection, which utilizes semantic matching with prototypes derived from ID class names. 
By harnessing the highly distinguishable representations of LLMs, we construct semantic prototypes for each ID class using a diversity-grounded prompt tuning approach.
We rigorously test our framework in a challenging OOD context, where ID and OOD classes are semantically close yet distinct, referred to as \emph{near} OOD detection. For a thorough assessment, we benchmark our method against the prevalent fine-tuning approaches. The experimental findings reveal that our method demonstrates superior performance in both few-shot ID intent classification and near-OOD intent detection tasks.

\end{abstract}

\section{Introduction}

Task-oriented conversational systems have become widespread across numerous scenarios, including banking, travel, and medical diagnosis~\cite{yan2017building,zhang2020recent, feng2024continual}, where they provide a variety of services. Intent recognition~\cite{zhang2022fine}, a key element of these systems, is crucial for enabling automated assistance to address customer needs. Given the wide range of user inquiries, out-of-distribution (OOD) intent detection~\cite{hendrycks2020pretrained, zhan2021out, uppaal2023fine, zhan2024vi} seeks to protect the intent recognition system by identifying and alerting to malformed inputs.

Recent advancements in large language models (LLMs)~\cite{touvron2023llama,chiang2023vicuna,openai2023gpt4, feng2023towards} have significantly improved the detection of semantically distinct out-of-distribution (far-OOD) intents~\cite{liu2023good}. However, identifying semantically similar (near-OOD) intents continues to pose significant challenges~\cite{fort2021exploring, liu2023good}. Given its critical importance in practical applications, our study concentrates on near-OOD detection, specifically in a few-shot scenario~\cite{jiang2024few,zhang2022pre}.
Particularly, in the few-shot learning context, the training model can receive only a few in-distribution (ID) examples per class.  Obtaining discriminative information from limited ID examples for OOD detection is inherently challenging~\cite{zhang2020discriminative}. Previous strategies have involved using extra unlabeled data~\cite{wang2023app} or generating pseudo-ID examples~\cite{zhan2022closer} to increase training samples. Instead, we focus on leveraging the inner knowledge of LLMs as a supplement without dataset expansion.

\begin{figure*}[t]
    \centering
    \includegraphics[width=0.95\linewidth]{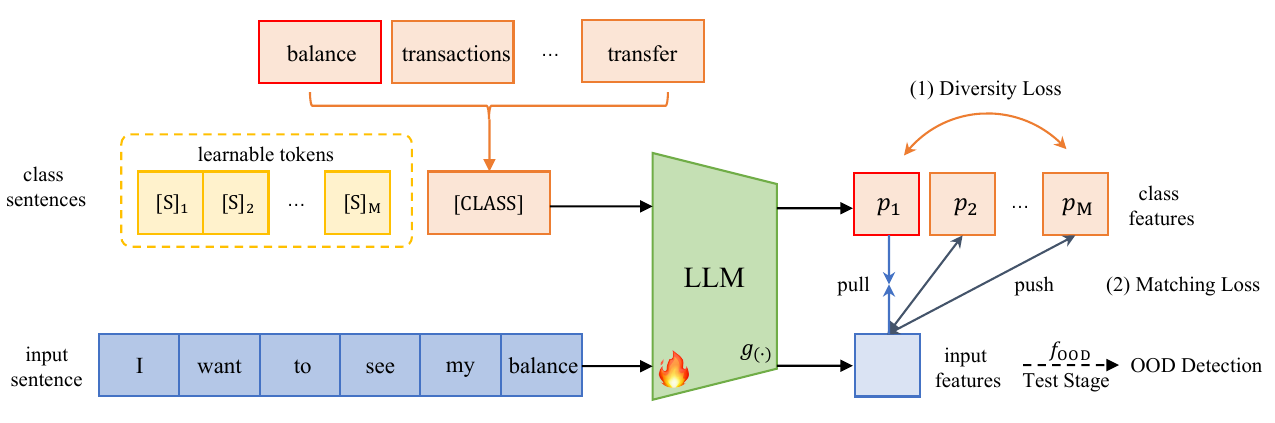}
    \caption{Our proposed semantic matching framework. We prompt class name into a sequence of learnable tokens and forward them into LLMs to generate class prototypes. With further training between prototypes and input representations via matching loss and diversity loss, better ID classification and OOD detection can be performed.}
    \label{fig:framework}
\end{figure*}

We propose a novel channel prototypical learning framework for few-shot near-OOD detection, shown in Figure~\ref{fig:framework}. 
It employs a set of learned class prototypes to conduct semantic matching for ID classification and OOD detection. To effectively leverage the pre-existing knowledge in LLMs and the limited ID examples, our framework involves using both the ID class names and utterances to develop the class prototypes. Drawing inspiration from channel models~\cite{min2021noisy, brown1993mathematics} and prompt tuning~\cite{lester2021power}, 
we feed into the LLMs the ID class names preceded by a series of learnable continuous prompt embeddings to generate a semantic prototype for each ID class.
Furthermore, to ensure a wider variety of class representations—a factor known to improve OOD detection~\cite{winkens2020contrastive}—we adopt a diversified learning strategy aimed at reducing the mutual information between classes.

For evaluation, we construct few-shot near-OOD intent detection tasks by sampling classes from the same domains in CLINC as ID and OOD. Since our method is a kind of fine-tuning framework, we compare it with widely used fine-tuning methods: generative tuning~\cite{liu2023good} and discriminative tuning\footnote{We use \texttt{LlamaForSequenceClassification} provided by Huggingface~\cite{wolf2019huggingface}}. Overall, our diversity-grounded channel prototypical learning approach can bring better performance compared to these two paradigms.

\section{Approach}

\paragraph{Problem Statement} A complete intent recognition system typically encompasses two tasks: ID intent classification and OOD intent detection. Formally, given an ID training set $ \{x_i\}_{i=1}^{N} \subset \mathcal{X}_{\text{ID}}$ with $N$ samples, an ID classifier $f_\text{ID}$ is trained, mapping each utterance into ID intent label set 
$\{y_j\}_{j=1}^{K} \subset \mathcal{Y}_{\text{ID}}$ with $K$ categories. In practical application, due to a possible distribution mismatch between the practical and training data, the ID classifier $f_\text{ID}$ may meet OOD samples ($x_j \notin \mathcal{X}_{\text{ID}}$). Therefore, an OOD confidence scoring function $f_\text{OOD}$ is applied to accept or reject such inputs.

\begin{table*}[t]
\centering
\resizebox{\linewidth}{!}{
\begin{tabular}{cccccccccc}
\hline
\hline

& &\multicolumn{4}{c}{\bf CLINC-Banking} & \multicolumn{4}{c}{\bf CLINC-Travel} \\
 \cmidrule(lr){3-6} \cmidrule(lr){7-10} 
Shot & Tuning Method & ID ACC$ \uparrow$ & AUROC  $\uparrow$  & FAR@95  $\downarrow$& AUPR $\uparrow$ & ID ACC$ \uparrow$ & AUROC  $\uparrow$  & FAR@95 $\downarrow$ & AUPR $\uparrow$ \\
\specialrule{0.05em}{0.3em}{0.3em}

  \multirow{3}{*}{5} & Generative$\dag$ & 0.882 & 0.962 & 0.255 &0.968 & 0.964 & 0.974 & 0.148 & 0.983\\
& Discriminative & 0.919 & 0.924 & 0.390 & 0.939 & 0.981 & 0.967 & 0.176 & 0.976\\
& Semantic Matching & \textbf{0.964} & \textbf{0.970} & \textbf{0.215} & \textbf{0.970} & \textbf{0.990} & \textbf{0.981} & \textbf{0.103} & \textbf{0.989}\\
 
 \cmidrule(lr){1-10}
 
\multirow{3}{*}{10} & Generative$\dag$ & 0.949 & 0.968 & 0.157 & \textbf{0.974} & 0.984 & 0.982 & 0.078 & 0.988 \\
& Discriminative & 0.959 & 0.955 & 0.266 & 0.966 & 0.991 & \textbf{0.984} & 0.075 & 0.989\\
& Semantic Matching & \textbf{0.967} & \textbf{0.970} & \textbf{0.149} & \textbf{0.974} & \textbf{0.993} &  \textbf{0.984} & \textbf{0.063} & \textbf{0.993} \\
 
 \cmidrule(lr){1-10}
 
\multirow{3}{*}{Full} & Generative$\dag$ & 0.973 &  0.964 & 0.147 &0.970 & 0.991 & 0.978 & 0.049 & 0.987\\
& Discriminative & \textbf{0.988} & 0.964 & 0.197 & 0.969 & 0.996 & \textbf{0.996} & 0.013 & \textbf{0.997}\\
& Semantic Matching & 0.983 & \textbf{0.971} & \textbf{0.141} & \textbf{0.980} & \textbf{0.997} & 0.994 & \textbf{0.006} &  0.996 \\

\hline
\hline
\end{tabular}
}
\caption{
The performance of LLaMA-7B fine-tuned with different methods for OOD detection and ID classification. ``Shot'' denotes the number of examples in the ID training and validation set.  We report the average results of five seeds. $\dag$ is cited from the original paper. AUROC, FAR@95, and AUPR are metrics for OOD detection.
}
\label{tab:fine_tune_llama}
\end{table*}

\subsection{Semantic Matching as ID Classification}\label{sec:sm}
Inspired by ~\citet{liu-etal-2024-good}, LLMs (like LLaMA~\cite{touvron2023llama}) have shown impressive isotropy~\cite{ethayarajh2019contextual,gao2019representation}, whereby the sentence embeddings produced by LLMs are distinguishable by Cosine distance. As such, we transform the ID classification task into a semantic matching task, shown in Figure~\ref{fig:framework}. 

\subsubsection{Diversity-grounded Channel Prototypical Learning}
\paragraph{Class Prototypes} Our key idea is to push the input sentence closer to its corresponding class prototypes (treated as class-center representations~\cite{snell2017prototypical}). Thus, we first prompt each class into a learnable sequence $\{\bm{c_i}\}_{i=1}^{K}$ based on its category name~\cite{min2021noisy} and then generate prototypes $\{\bm{p_i}\}_{i=1}^{K}$ with LLM model $g(\cdot)$ through:
\begin{subequations}
\begin{equation} \label{eq:cls}
    \bm{c_i} = [\text{S}]_1[\text{S}]_2...[\text{S}]_M[\text{NAME}],
\end{equation}
\begin{equation} \label{eq:cls}
    \bm{p_i} = g(\bm{c_i}),
\end{equation}
\end{subequations}
where $[\text{S}]_j$ ($j \in \{1,...,M\}$) is a learnable token with the same dimension as word embeddings (i.e., 4096 for LLaMA), M is the number of learnable tokens, and ``NAME'' corresponds to token embeddings of category names. In practice, the initialization of each learnable token $[\text{S}]_j$ is derived from the token embeddings of a specific prompt, e.g., ``[SCENARIO] intent of'' where SCENARIO is ``banking'' for a banking system and is ``travel'' for a travel system.

\paragraph{Diversity Learning} To employ semantic matching as classification, the semantics between class prototypes $\bm{p_i}$ must be distinguishable with satisfied mutual exclusion. Therefore, a diversity loss is proposed to help the class prototypes focus on diversified independent semantics:
\begin{equation} 
    \mathcal{L}_{\text{Diversity}} = \frac{1}{K^2} \sum_{i=1}^K \sum_{j=1, j \neq i}^K 
    (\text{cos}(\bm{p_i},\bm{p_j}))^2,
\end{equation}
where $\text{cos}$ denotes the Cosine similarity. Obviously, the semantic similarities between different class prototypes are expected to be $0$.

\subsubsection{Semantic Matching} 
In representation space, for input sentence $\bm{x}$, we push its representation closer to the corresponding class prototype $\bm{p_j}$ and farther away from other ones in a contrastive manner: 
\begin{equation} \label{eq:match}
    \mathcal{L}_{\text{Match}} = -\log \frac{\text{exp}(\text{cos}(g(\bm{x}),\bm{p_j})/\tau)}{\sum_{i=1}^K \text{exp} (\text{cos}(g(\bm{x}), \bm{p_i})/\tau)},
\end{equation}
where $\tau$ is the temperature~\cite{hinton2015distilling}.

\subsubsection{Optimization}
We optimize the LLM model $g(\cdot)$ as well as learnable tokens $\{[S]_j\}_{j=1}^{M}$ with a joint training objective:
\begin{equation} \label{eq:cls}
    \mathcal{L}_{\text{Joint}} = \lambda \mathcal{L}_{\text{Diversity}} + 
    \mathcal{L}_{\text{Match}} ,
\end{equation}
where $\lambda$ is a balancing hyper-parameter. Note that different from ~\citet{lester2021power} that froze the backbone LLM, we utilize the low-rank adaptation (LoRA)~\cite{hu2021lora} technique to parameter-efficient tune it (detailed in Section~\ref{sec:exp_setup}).

\subsection{OOD Detection with Post-hoc Scoring} 
After fine-tuning with ID data, we mainly focus on a distance-base scoring function in post-hoc paradigm, i.e., Cosine distance (Cosine)~\cite{zhou2021contrastive}, due to its exceptional performance in the LLM era~\cite{liu2023good}. In particular, the Cosine confidence score is defined as the highest cosine similarity between the test input representation and the representations in the validation set:
    \begin{equation}
        \mathcal{S}(x) = \max \{\text{cos}(z, z_{i}^{val})\}_{i=1}^{V},
    \end{equation} 
where representation $z = g (x) $ is the output obtained from the penultimate layer of LLMs.

\section{Experiment}
\subsection{Experimental Setup}\label{sec:exp_setup}

\paragraph{Datasets}  We explore a more challenging scenario, termed the \emph{near-}OOD setting, where ID and OOD samples come from the same domain but with disjoint label sets. Following ~\citet{liu2023good}, we use \texttt{CLINC150}~\cite{larson2019evaluation} intent dataset and choose \emph{Banking} and \emph{Travel} domains.  Within each domain, 50\% of the classes are designated as ID, and the remaining classes as OOD. Detailed dataset statistics are provided in Table~\ref{tab:data}.

\begin{table}[t]
\centering
\resizebox{\linewidth}{!}{
\begin{tabular}{ccccc}
\hline

Datasets & Classes & Training & Validation & Test\\
\hline
 \emph{CLINC150-Banking} & 15 &1500 & 300 & 450  \\
 \emph{CLINC150-Travel}  & 15 &1500 & 300 & 450 \\

\hline
\end{tabular}
}
\caption{
Data composition of each dataset.
}
\label{tab:data}
\end{table}



\paragraph{Evaluation Metrics} In accordance with ~\citet{liu2023good}, we adopt three widely accepted measures for OOD detection: (1) AUROC (area under the receiver operating characteristic curve), (2) FAR@95 (false alarm rate at 95\% recall), which indicates the likelihood of erroneously labeling a negative sample as positive when recall or true positive rate (TPR) is 95\%; Here, we consider the the OOD class as negative, and (3) AUPR (area under the precision-recall curve). Furthermore, accuracy serves as a metric for the ID classification task.

\paragraph{Implementation Details} Following prior work for OOD detection with LLMs~\cite{liu2023good}, we employ the LLaMA-7B as $g(\cdot)$.
For both class prototypes and input representations, we use the representation of the last token. $\tau$ in Eq.~\ref{eq:match} is set to 0.01 and $\lambda$ in Eq.~\ref{eq:cls} is set to 0.2 through cross-validation. We train the whole network for 25 epochs, and the model that achieved the lowest loss on the validation set will be used for testing. We use AdamW optimizer with a learning rate of $1 \times 10^{-4}$, further decayed by linear schedule.  For discriminative tuning, we use the same configurations. More details can be found in Appendix~\ref{sec:appendix}.

\subsection{Main Results}
The main results are shown in Table~\ref{tab:fine_tune_llama}. On one hand, the proposed semantic matching fine-tuning framework consistently outperforms other tuning methods in most cases, especially for the intent classification task with low training data. Take the results of 5-shot classification on CLINC-Banking as an example. Semantic matching outperforms discriminative and generative tuning by 4.5\% and 8.2\% respectively, demonstrating the high data-efficient ability of our framework. Meanwhile, the OOD detection performance of Cosine distance improves accordingly due to more accurate and dense representations of classes. On the other hand, in the full-shot setting, while semantic matching yields slightly lower ID classification results on CLINC-Banking compared to the discriminative one, its OOD detection performance is superior, and it also remains competitive on the CLINC-Travel dataset.

\subsection{Analysis and Ablation Study}

\begin{figure}[t]
    \centering
    \includegraphics[width=1\linewidth]{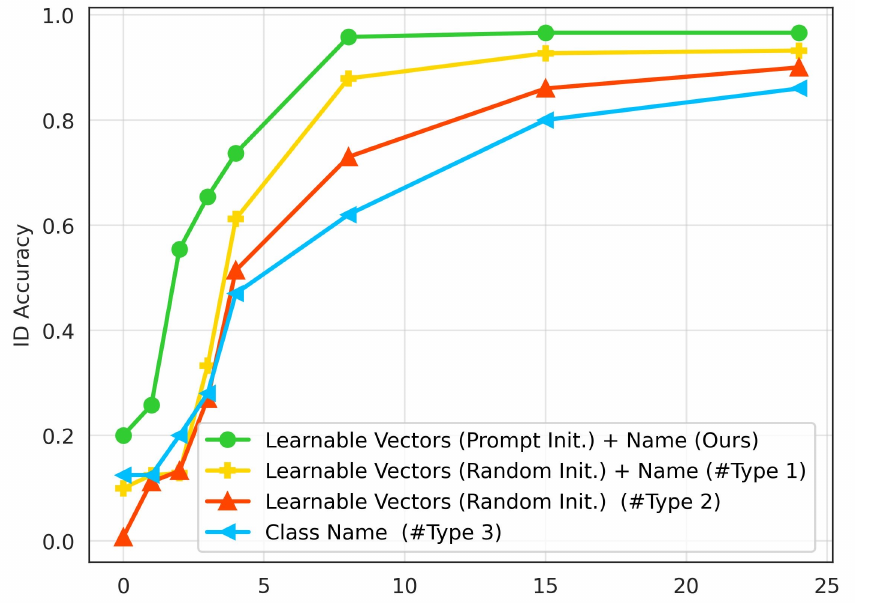}
    \caption{Performance of various class prototypes in the 5-shot scenario using CLINC-Banking dataset.}
    \label{fig:variants}
\end{figure}
\paragraph{Variants of Class Prototypes} As elaborated in Sec~\ref{sec:sm}, we use learnable vectors initialized with scenario-specific prompts plus a class name embedding to generate prototypes. A question worth exploring is what are the effects of other choices? Here, we additionally try 3 types: (1) randomly initialized learnable vectors plus class name embeddings; (2) only randomly initialized learnable vectors; (3) only class name embeddings, e.g., ``transactions'' in the CLINC-Banking dataset.  Performance changes for ID intent classification as training progresses are shown in Figure~\ref{fig:variants}. It reveals that (1) Our design (green curve) not only obtains the best performance but only requires a few iterations to achieve optimally. (2) Overall, learnable tokens are very important and can bring better results than fixed embeddings of class names (Type 3).

\begin{figure}[t]
    \centering
    \includegraphics[width=1\linewidth]{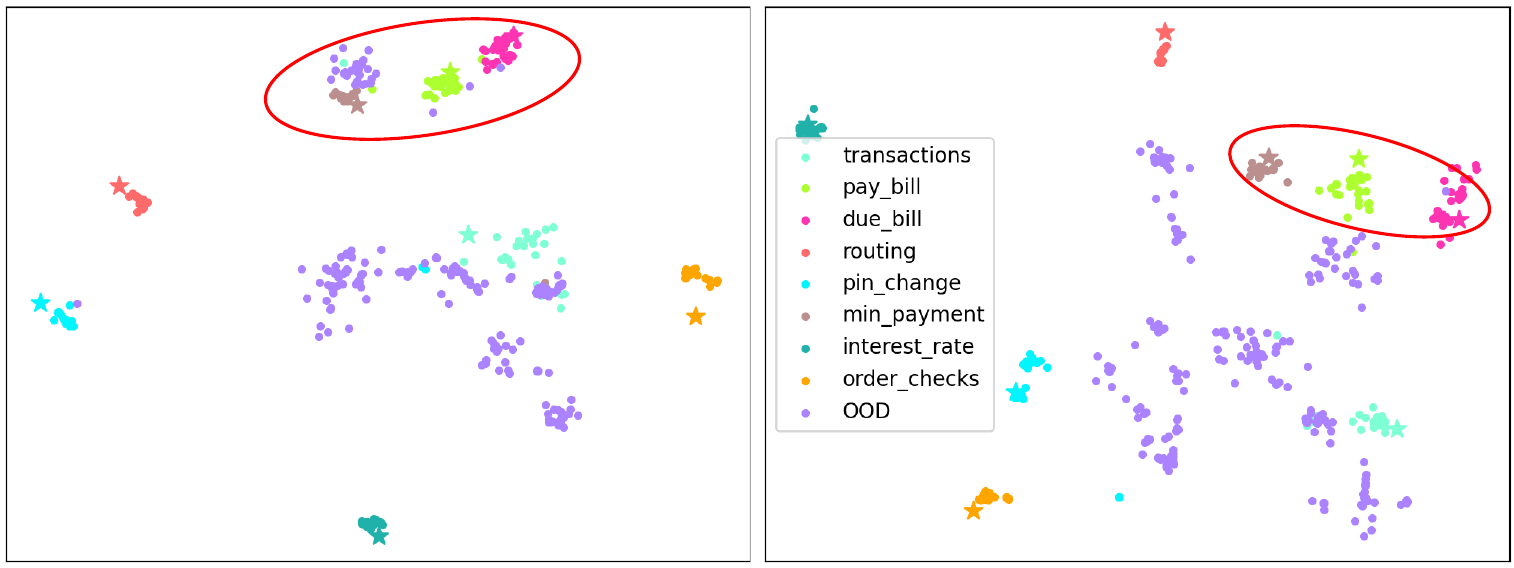}
    \caption{UMAP~\cite{mcinnes2018umap} visualization of representations of test set and OOD data from 5-shot CLINC-Bank. The purple means the OOD data. The star indicates the learned prototype of each class.}
    \label{fig:embedding}
\end{figure}

\begin{table}[h]
\centering
\resizebox{\linewidth}{!}{
\begin{tabular}{ccccc}
\hline

&\multicolumn{4}{c}{\bf CLINC-Banking (5-shot)} \\

Tuning Method & ID ACC$ \uparrow$ & AUROC  $\uparrow$  & FAR@95  $\downarrow$& AUPR $\uparrow$ \\
\hline
Generative & 0.882 & 0.962 & 0.255 &0.968 \\
Discriminative & 0.919 & 0.924 &  0.390 & 0.939 \\
\hline
 only $\mathcal{L}_{\text{Match}}$ & 0.943 & 0.950 & 0.324 & 0.961 \\
 $\mathcal{L}_{\text{Match}}$ + $\mathcal{L}_{\text{Diversity}}$  & \textbf{0.964} & \textbf{0.970} & \textbf{0.215} & \textbf{0.970}\\

\hline
\end{tabular}
}
\caption{
Effectiveness of Diversity Learning.
}
\label{tab:ablation}
\end{table}

\paragraph{Effectiveness of Diversity Learning} We investigate the effectiveness of each training objective in the semantic matching framework with results shown in Table~\ref{tab:ablation}. We can find that (1) Incorporating diversity learning can boost the overall performance on both tasks. Notably, it enhances about 11\% on the FAR@95 metric, showing strong effectiveness; (2) To better understand why diversity learning provides an improvement, we visualize the corresponding sentence representations in Figure~\ref{fig:embedding}. Intuitively, diversity learning makes the separation between classes clearer, thereby enhancing the classification performance of ID data. Simultaneously, it could implicitly distinguish potential OOD data, clarifying the boundaries between ID and OOD data and thus improving OOD detection accuracy.

\section{Conclusion}
In this paper, we have proposed a semantic matching fine-tuning framework for boosting ID intent classification and OOD intent detection. By prompting class names into learnable vectors and pushing their representations yielded by LLMs closer to belonged sentence representations, semantics-awareness classification has been conducted. Extensive experiments compared to generative and discriminative tuning, as well as detailed analyses, were presented, demonstrating the effectiveness and data efficiency of our proposal.

\section*{Limitation}
This paper mainly has two limitations: 
(1) we only utilized the LLaMA-7B model, without investigating other open-source large models such as OPT~\cite{zhang2022opt}, LLaMA-2\&3~\cite{touvron2023llama}, and Mistral~\cite{jiang2023mistral}. Furthermore, the impact of instruct-tuning LLMs like LLaMA-Chat on this task was also left unexplored. In the proposed semantic matching framework, strong backbone models could potentially yield superior overall results.
(2) 
Presently, this study predominantly builds upon prior research by ~\citet{liu2023good}, focusing on the CLINC dataset. Therefore, intent datasets such as BANKING77~\cite{casanueva2020efficient} and StackOverflow~\cite{xu2015short} remain unexplored in this context. Further exploration of these datasets would be beneficial.



\bibliography{custom}

\appendix
\section{Appendix}
\label{sec:appendix}
\subsection{Training Details}
All experiments are conducted over five seeds (1, 2, 3, 4, 5) on an NVIDIA A100 80G GPU card. The LoRA configurations are that rank $r$ is 16, scaling $\alpha$ is 16, and query/key/value/output projection matrices $\{ W_q, W_k, W_v, W_o\}$ in each self-attention module need to be updated. 


\end{document}